\documentclass{article}
\usepackage{graphicx}
\usepackage{amsmath}
\usepackage{amssymb}
\usepackage{booktabs}
\usepackage{multirow}
\usepackage[utf8]{inputenc}
\usepackage{hyperref}

\begin{document}

\title{Detecting Deepfakes with Multivariate Soft Blending and CLIP-based Image-Text Alignment}

\author{Jingwei Li \\
Zhejiang Gongshang University \\
25090246@mail.zjgsu.edu
\and
Jiaxin Tong\\
Zhejiang Gongshang University \\
jiaxin.tong@mail.zjgsu.edu
\and
Pengfei Wu\\
Zhejiang Gongshang University \\
pengfei.wu1@mail.zjgsu.edu
}

\date{}

\maketitle

\begin{abstract}
With the rapid advancement of facial image synthesis technology, the creation of highly realistic fake facial images and videos has become increasingly accessible, posing significant threats to social information security. However, existing deepfake detection methods suffer from insufficient accuracy and poor generalization due to the substantial distributional differences among samples generated by various forgery techniques. To address these challenges, this paper proposes a novel multivariate and soft blending sample-driven image-text alignment method for deepfake detection, which leverages the multimodal alignment of images and text to capture subtle traces of facial forgery. Recognizing that traditional face forgery detection methods are typically trained on single-mode forged images and struggle with complex forgery patterns, we introduce a Multivariate and Soft Blending Augmentation (MSBA) strategy. This data augmentation technique generates images by randomly blending forged images of different modes with various weights, forcing the network to simultaneously capture clues from multiple forgery patterns and thereby enhancing its detection capability for complex and unknown forgeries. Furthermore, to counteract performance degradation caused by the diverse forgery modes and intensities in facial images, we design a Multivariate Forgery Intensity Estimation (MFIE) module based on the MSBA strategy. This module guides the image encoder to learn generalized features tailored to different forgery modes and intensities, ultimately improving the overall detection accuracy of the framework. Experimental results demonstrate that in in-domain tests, our approach achieves improvements of 3.32\% in Accuracy (ACC) and 4.02\% in the Area Under the Curve (AUC) metric compared to the best-performing baseline method. In cross-domain tests, our method is evaluated against six state-of-the-art methods across five datasets, showing an average AUC improvement of 3.27\%. Ablation studies confirm the positive contributions of both the MSBA strategy and the MFIE module to face forgery detection performance. The CLIP-based network framework designed for this task significantly improves detection accuracy, with the proposed MSBA and MFIE modules playing crucial supporting roles in surpassing existing methods. However, due to the reliance on large-scale vision-language models, the proposed method has a relatively high number of parameters and computational complexity, which imposes limitations on its inference speed. Future work will focus on reducing the model's computational overhead while maintaining or even improving its accuracy and robustness in face forgery detection.
\end{abstract}

\section{Introduction}
The rapid evolution of deep learning-based facial forgery techniques, progressing from simple image editing to sophisticated generative adversarial networks (GANs), has enabled the creation of highly realistic fake facial images and videos with alarming ease. As facial biometrics become increasingly integral to identity verification in critical domains such as access control, secure payments, financial services, and social media, the malicious use of forged identities poses severe threats to personal privacy, financial security, and societal trust. Consequently, the development of robust and generalizable face forgery detection methods is urgently required to mitigate these pervasive risks.

In response to this challenge, a significant body of research has emerged \cite{chen2024translationfusionimproveszeroshot}. From a data-centric perspective, given the constant proliferation of new forgery methods which makes exhaustive data collection impractical, some studies focus on synthesizing diverse forgery samples. For instance, Li \emph{et al.} \cite{li2024longcontextllmsstrugglelong} and Zhao \emph{et al.} \cite{zhao2024harmonizing} generated blended images (BI) by mixing two different facial images with smoothed boundaries. Shiohara and Yamasaki \cite{teknium2024hermes3technicalreport} proposed the Self-Blended Image (SBI) method, which applies subtle color and affine transformations to facial regions before blending, thereby encouraging models to learn more generalizable features. From a model architecture perspective, the substantial distributional shift between samples generated by different forgery algorithms often leads to overfitting. This has spurred innovations in specialized network design. Some approaches exploit frequency domain anomalies, such as using Discrete Fourier Transform (DFT) \cite{achiam2023gpt} or Discrete Cosine Transform (DCT) \cite{veturi2024ragbasedquestionansweringcontextual} analyses. Others focus on noise patterns with Spatial Rich Model (SRM) filters \cite{lu2024bounding} or employ decoupling frameworks to isolate content from forgery traces \cite{lightrag}. Recent methods also explore multi-modal fusion \cite{zhang2023blind}, patch-based comparisons \cite{feng2023unidoc}, and strategies to uncover identity inconsistencies \cite{huang2025mindev, jinensibieke2024goodllmsrelationextraction}.

Despite these advances, existing methods still face two core limitations: insufficient detection accuracy and poor generalization capability. First, most detectors are trained on data from a limited set of single-mode forgeries, rendering them vulnerable to complex, blended, or novel attack methods unseen during training. Second, the vast diversity in forgery techniques and manipulation intensities presents a significant learning challenge, often degrading model robustness and leading to overfitting on dataset-specific artifacts.

Recently, large-scale pre-trained Vision-Language Models (VLMs) like CLIP (Contrastive Language-Image Pre-training) \cite{graphrag} have demonstrated remarkable success across various vision tasks \cite{liu2023spts}, owing to their powerful, aligned cross-modal representations. Their proven capability in recognizing diverse visual patterns aligns perfectly with the core requirement of forgery detection—identifying subtle, varied manipulation traces. Pioneering work has begun adapting such models for general deepfake detection \cite{Khan2024}, showcasing their potential. Inspired by this, our work is among the first to deeply integrate a VLM framework specifically for the face forgery detection task.

The principal contributions of this paper are threefold. First, we propose a novel multivariate and soft blending sample-driven image-text alignment network for face forgery detection. By leveraging the semantically aligned multimodal representations of CLIP, our framework significantly enhances detection accuracy and generalization. Second, to combat the limitation of single-mode training, we introduce a Multivariate and Soft Blending Augmentation (MSBA) strategy. This method synthesizes training samples containing complex, blended forgery patterns by randomly combining images from different forgery methods with soft weights, forcing the model to learn disentangled and concurrent forgery clues. Third, building upon MSBA, we design a Multivariate Forgery Intensity Estimation (MFIE) module. This module guides the image encoder to explicitly learn and estimate the intensity and composition of forgeries within an image, promoting the extraction of more generalized and informative features. Extensive experiments demonstrate that the synergistic effect of the MSBA strategy and the MFIE module enables our model to achieve state-of-the-art performance in both in-domain and challenging cross-domain evaluations, exhibiting superior robustness and generalization.

\section{Related Works}

\subsection{Traditional Blind Image Separation Methods}

The foundation of Blind Image Separation (BIS) traces back to classical Blind Source Separation (BSS) techniques, which were initially developed for signal processing applications. Early approaches predominantly relied on statistical methods and mathematical constraints to tackle the inherent ill-posed nature of separation problems. 

\textbf{Independent Component Analysis (ICA)} and its variants formed the cornerstone of early BIS research. The fundamental assumption of ICA revolves around the statistical independence and non-Gaussian distribution of source signals. While theoretically sound for simple signal mixtures, these methods encounter significant limitations when applied to complex image data. Real-world images often exhibit substantial feature correlations and distribution dependencies, violating the core independence assumption. Moreover, ICA-based approaches struggle with nonlinear mixing scenarios commonly encountered in practical imaging conditions, such as atmospheric interference, optical distortions, and sensor nonlinearities.

\textbf{Sparse coding and low-rank decomposition} methods emerged as subsequent improvements, introducing additional prior constraints to enhance separation performance. These techniques leverage the inherent sparsity of image representations in certain domains or the low-rank structure of image components. However, they remain inadequate for characterizing the complex, multi-scale textures and structures present in natural images. The hand-crafted priors often fail to capture the rich statistical regularities of real-world image distributions, particularly when dealing with highly correlated sources or severe mixing conditions.

\subsection{Deep Learning-Based Approaches}

The advent of deep learning marked a paradigm shift in BIS research, enabling data-driven learning of complex feature representations and separation mappings.

\subsubsection{CNN-Based Architectures}
Convolutional Neural Networks brought revolutionary advances through their hierarchical feature extraction capabilities. The local receptive fields of CNNs enable efficient learning of spatial correlations, making them particularly suitable for image separation tasks. Early CNN-based BIS methods focused on encoder-decoder architectures that learn direct mappings from mixed images to separated sources. However, these approaches often suffer from limited receptive fields, struggling to capture long-range dependencies crucial for global separation consistency.

\subsubsection{GAN-Based Frameworks}
Generative Adversarial Networks introduced a powerful alternative by circumventing explicit prior assumptions through adversarial training. The seminal work by Hoshen et al. \cite{Ho2020} pioneered the application of CycleGAN for unsupervised single-channel BIS, establishing bidirectional mappings between mixed and source domains. This approach significantly advanced the field by reducing dependency on paired training data. Subsequent improvements incorporated attention mechanisms (Attention-GAN) and transformer architectures (Transformer-GAN) to enhance global context modeling. Sun et al. \cite{sun2025attentive} integrated self-attention mechanisms with GANs to capture global dependencies, while Su et al. \cite{sun2025attentive} designed Transformer-based GANs to balance local detail preservation with global feature extraction.

\subsubsection{Advanced Deep Learning Variants}
Recent years have witnessed increasingly sophisticated architectures addressing specific BIS challenges. Liu et al. \cite{liu2025gated} developed parallel twin adversarial networks to establish relationships between mixed signals and multiple sources. Jia et al. \cite{jia2025meml} addressed data scarcity through cascaded UGAN-PAGAN frameworks that generate synthetic training samples while performing separation. Despite these advancements, deep learning methods continue to face challenges in real complex scenarios, including feature distribution uncertainty, nonlinear mixing complexities, and irregular noise interference.

\subsection{Diffusion Models in Image Processing}

Diffusion Models have recently emerged as state-of-the-art generative frameworks, demonstrating remarkable capabilities across diverse image processing tasks.

\subsubsection{Foundational Diffusion Architectures}
The denoising diffusion probabilistic models (DDPM) proposed by Ho et al. \cite{Ho2020} established the fundamental framework by combining discrete-time Markov chains with variational inference. This approach progressively transforms data through forward diffusion and learns reverse processes for high-quality generation. Subsequent improvements include score-based generative modeling and stochastic differential equation formulations that provide more flexible and efficient sampling strategies.

\subsubsection{Applications in Image Restoration}
Diffusion models have shown exceptional performance in various image restoration tasks. Saharia et al. \cite{wu2024medicalgraphragsafe} developed SR3 for image super-resolution through iterative refinement, while Liu et al. \cite{li2025audio} proposed Diffusion-based Plugin frameworks for diverse low-level vision tasks. Recent applications include image denoising \cite{es2023ragasautomatedevaluationretrieval, chen2024translationfusionimproveszeroshot}, super-resolution \cite{zhang2023blind, zhao2025tabpedia}, and image editing \cite{huang2025mindev, wang2025systematic}. These successes demonstrate the versatility and power of diffusion processes in handling complex image transformations.

\subsubsection{Specialized Diffusion Variants}
Several specialized diffusion architectures have been developed for specific challenges. Wang et al. \cite{wang2025pargo} integrated fuzzy logic with diffusion models for document image restoration, enhancing robustness to random factors. He et al. \cite{gunther2024late} proposed Bayesian uncertainty-guided diffusion models (BUFF) that incorporate high-resolution uncertainty masks to reduce artifacts in complex texture regions. Chen et al. \cite{zhang2023blind} designed controllable sampling strategies that restrict potential output ranges through initial sample prediction, improving deblurring accuracy.

\paragraph{Foundational Models and OCR Toolkits.}

The capabilities of large language models (LLMs) and large multimodal models (LMMs) form the bedrock of modern document understanding. Models such as GPT-4 \cite{achiam2023gpt} and the LLaMA family \cite{llama} have demonstrated exceptional reasoning and generative abilities, setting new standards for zero-shot and few-shot learning across modalities. Specialized instruction-tuned models like Hermes 3 \cite{teknium2024hermes3technicalreport} further optimize these capabilities for interactive tasks. Concurrently, the field of optical character recognition (OCR) has matured, with industrial-grade toolkits like PaddleOCR \cite{paddleocr2023} and the venerable Tesseract \cite{Tesseract} providing robust, multilingual text extraction. These tools often serve as the critical first step in many document processing pipelines, translating raw visual data into textual tokens for higher-level analysis.

\paragraph{End-to-End Text Detection and Recognition.}

Moving beyond traditional OCR pipelines, research has focused on unifying text detection and recognition into end-to-end frameworks. Pioneering works such as \cite{tang2022few} introduced efficient feature sampling and grouping strategies for robust scene text detection. \cite{tang2022optimal} further enhanced recognition accuracy through reinforcement learning to adjust bounding boxes. This line of research culminated in models like SPTS v2 \cite{liu2023spts}, which achieved state-of-the-art performance in end-to-end text spotting. Subsequent works like that of \cite{zhao2024multi} introduced in-context learning mechanisms to create adaptive, ego-evolving recognizers. The work by \cite{tang2023character} expanded the scope to real-world applications, organizing a large-scale competition for character recognition in street view shop signs.

\paragraph{Document Structure Parsing and Layout Analysis.}

To comprehend complex documents, it is essential to parse their structural elements beyond mere text. Recent efforts have concentrated on building unified, multimodal models for document understanding. \cite{feng2023unidoc} proposed a universal large multimodal model capable of simultaneous text detection, recognition, spotting, and understanding. Its successor, \cite{feng2024docpedia}, further leveraged frequency domain analysis for versatile document parsing. A significant methodological shift is represented by \cite{lu2024bounding}, which interleaves layout information (bounding boxes) directly with text tokens within an LLM, enabling the model to reason about spatial relationships. This vision is further advanced by comprehensive frameworks like \cite{feng2025dolphin}, which employs heterogeneous anchor prompting for detailed document image parsing. For complex engineering drawings, specialized approaches have been developed for information extraction, such as those focusing on geometric dimensions and tolerances \cite{Gao2005, Khan2024, Xu2024}.

\paragraph{Text-Centric Multimodal Understanding and Reasoning.}
Recognizing that deep understanding requires reasoning over the \emph{content} of visual text, the research frontier has shifted towards text-centric multimodal learning. \cite{tang2024textsquare} demonstrated the power of scaling up visual instruction tuning specifically focused on text, achieving superior performance across a wide array of tasks. To rigorously evaluate such capabilities, several benchmarks have been introduced. \cite{tang2024mtvqa} established MTVQA to assess multilingual text-centric visual question answering, while \cite{shan2024mctbench} proposed MCTBench to test multimodal cognition towards text-rich scenes. These benchmarks challenge models to not only read but also comprehend and reason based on textual information embedded in images. Further extending the reasoning paradigm, \cite{lu2025prolonged} proposed an adaptive routing mechanism to enhance the efficiency of LLM/MLLM reasoning processes. Complementary research explores the integration of other sensory data, such as audio-visual cues for real-world tasks like idling vehicle detection \cite{li2025audio}. The challenge of understanding long-form documents is addressed by works like \cite{liu2025resolving}, which employs agentic context engineering to manage evidence sparsity.

\paragraph{Retrieval-Augmented Generation for Enhanced Knowledge Grounding.}
To mitigate hallucinations and ground responses in external knowledge, Retrieval-Augmented Generation (RAG) has emerged as a pivotal technique. Pioneering work by \cite{lewis2020retrieval} demonstrated its effectiveness for knowledge-intensive NLP tasks. The survey by \cite{gao2023retrieval} provides a comprehensive overview of this rapidly evolving paradigm. To address the challenge of evaluating RAG systems without ground-truth annotations, \cite{es2023ragasautomatedevaluationretrieval} introduced the RAGAS framework. Recent innovations focus on structuring the knowledge base, such as graph-based RAG (GraphRAG) \cite{graphrag} for holistic insights, with applications like MedGraphRAG for the medical domain \cite{wu2024medicalgraphragsafe} and CommunityKG-RAG for fact-checking \cite{chang2024communitykgragleveragingcommunitystructures}. Efforts to improve efficiency include prompt compression techniques like LongLLMLingua \cite{jiang2024longllmlinguaacceleratingenhancingllms} and simplified, fast retrieval frameworks like LightRAG \cite{lightrag}. Notably, Anthropic's introduction of Contextual Retrieval \cite{contextual} highlights a method to preserve context during the encoding process, significantly improving retrieval accuracy for RAG systems.

\paragraph{Cross-lingual and Low-Resource Information Extraction.}
The application of these technologies to low-resource languages (LRLs) presents unique challenges. Research in this area investigates the performance of LLMs on tasks like relation extraction \cite{jinensibieke2024goodllmsrelationextraction} and explores settings for their effective deployment in multilingual, multimodal, and dialectal contexts \cite{Alam2024LLMsFL}. Frameworks like TransFusion \cite{chen2024translationfusionimproveszeroshot} leverage translation and annotation fusion to improve zero-shot cross-lingual information extraction, aiming to close the performance gap between high and low-resource languages.

\paragraph{Synthesis and Motivation.}
In summary, the field has progressed from isolated, sequential components (OCR $\rightarrow$ NLP) towards integrated, parallel-processing multimodal systems. While existing approaches in document parsing and structured RAG offer powerful tools for document understanding and knowledge grounding, they often remain as separate modules or rely on upstream OCR, which can be brittle. There is a clear opportunity for a more holistic framework that unifies fine-grained visual perception, layout-aware reasoning, and context-sensitive retrieval into a single, end-to-end model. Our work is motivated by this gap, seeking to develop a system that treats the document itself as a queryable knowledge graph, enabling direct, robust, and interpretable reasoning over complex, real-world documents without the error-propagation pitfalls of traditional pipelines.

\begin{figure}[t]
    \centering
    \includegraphics[width=0.95\linewidth]{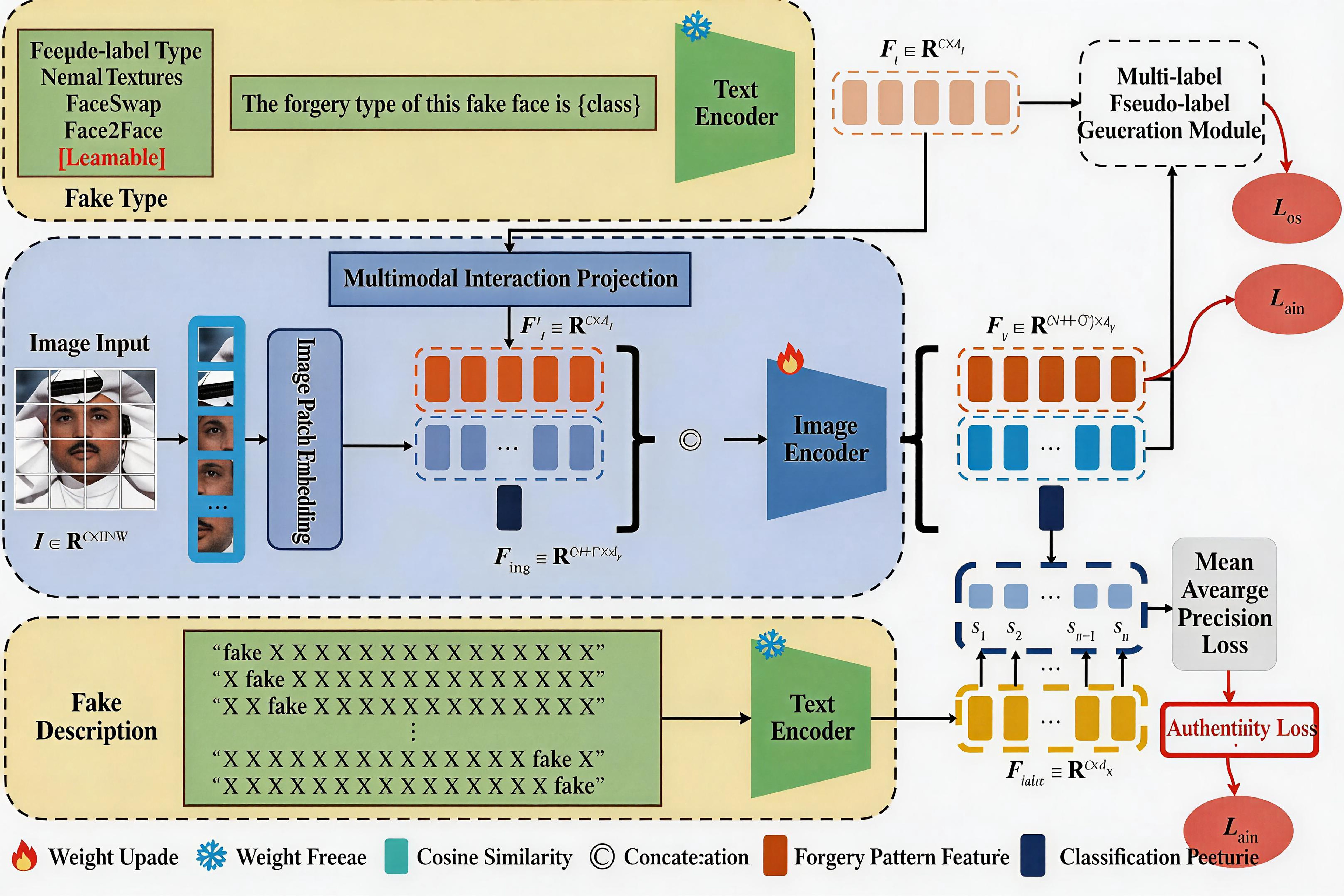} % Replace with actual figure path
    \caption{Schematic diagram of the proposed MSBA-CLIP framework. The input image undergoes MSBA augmentation. The CLIP-based encoder fuses image and text prompts. The Multivariate Forgery Intensity Estimation (MFIE) module predicts both the spatial intensity map and the method-specific blending weights. The final prediction is a fusion of a classification head output and a text-image similarity score.}
    \label{fig:framework}
\end{figure}

\section{Method}
\label{sec:method}

Our proposed method, named \textbf{Multivariate and Soft Blending Augmentation with CLIP-guided Forgery Intensity Estimation (MSBA-CLIP)}, is designed to tackle the limitations of existing face forgery detection methods by integrating multimodal learning, advanced data augmentation, and explicit forgery intensity modeling. The overall architecture, depicted in Figure \ref{fig:framework}, leverages the power of Contrastive Language-Image Pre-training (CLIP) \cite{tang2023character} as a backbone while introducing novel components specifically tailored for the face forgery detection task.

\subsection{Overall Architecture: Image-Text Alignment Framework}
\label{subsec:overall_arch}

Our network architecture is built upon the CLIP-ViT model \cite{Radford2021Learning}. Unlike conventional methods that rely solely on visual features, our framework explicitly integrates text modality to guide the visual feature extraction process towards more generalizable forgery traces. This is motivated by recent successes in adapting vision-language models for complex visual understanding tasks \cite{lu2024bounding}.

Given an input face image $I \in \mathbb{R}^{H \times W \times 3}$, we first encode it into a sequence of visual tokens via CLIP's patch embedding layer, yielding $F_{\text{img}} \in \mathbb{R}^{(N+1) \times d_v}$, where $N = HW / P^2$ is the number of patches (with patch size $P=16$), and the extra token represents the global [CLS] token.

To inject semantic guidance, we construct a text prompt conditioned on the forgery type: \textit{``The forgery type of this fake face is \{class\}''}, denoted as $t_c$. Here, \textit{class} $\in \{\text{DeepFakes, NeuralTextures, FaceSwap, Face2Face, Unknown}\}$, representing the five categories from the FF++ dataset \cite{Rombach2022}. This prompt is encoded by CLIP's text encoder $G_t(\cdot)$ to obtain the text feature $F_t \in \mathbb{R}^{1 \times d_t}$.

A key innovation is the introduction of a \textbf{Multimodal Interaction Projection (MIP)} layer. This layer projects the text feature $F_t$ into the same dimensional space as the visual tokens:

\begin{equation}
    F_t' = \text{MLP}(\text{LayerNorm}(F_t)), \quad F_t' \in \mathbb{R}^{1 \times d_v}
\end{equation}
where MLP is a two-layer perceptron and LayerNorm denotes layer normalization \cite{Rombach2022}.

Subsequently, the projected text feature $F_t'$ is concatenated with the visual token sequence $F_{\text{img}}$ along the token dimension. This concatenated token sequence is then fed into the Transformer-based visual encoder $G_v(\cdot)$ of CLIP. The early fusion of text guidance encourages the visual encoder to attend to image regions that are semantically relevant to the described forgery type, fostering a more focused feature extraction.

For the final real/fake binary classification, we adopt a dual-supervision strategy. First, we use the standard classification head on the final [CLS] token representation $F_{\text{cls}}$. Second, to enhance textual diversity and robustness, we compute a semantic similarity score. We generate $L$ generic negative text descriptions $t_{\text{fake}}$ (e.g., \textit{``This is a fake face image.''}), encode them to get features $F_{\text{fake}} \in \mathbb{R}^{L \times d_t}$, and compute their average cosine similarity with $F_{\text{cls}}$:
\begin{equation}
    \tilde{y}_{\text{sim}} = \frac{1}{L} \sum_{i=1}^{L} \text{sim}\left(G_t(t_{\text{fake}}^{(i)}), \text{Proj}(F_{\text{cls}})\right)
\end{equation}
where $\text{sim}(\cdot)$ denotes cosine similarity and $\text{Proj}(\cdot)$ is a linear projection to align dimensions. The final prediction $\hat{y}$ is obtained by fusing the logits from the classification head and the similarity score.

\subsection{Multivariate and Soft Blending Augmentation (MSBA)}
\label{subsec:msba}

A fundamental challenge in deepfake detection is the model's tendency to overfit to artifacts specific to a single forgery method, leading to poor generalization. To mitigate this, we propose a novel \textbf{Multivariate and Soft Blending Augmentation (MSBA)} strategy. The core idea is to create synthetic training samples that contain blended patterns from multiple forgery methods, thereby forcing the model to learn disentangled and more generalizable forgery features. This concept draws inspiration from data augmentation techniques in computer vision \cite{chen2024translationfusionimproveszeroshot} and is adapted for the specific domain of facial forgeries.

\begin{figure}[t]
    \centering
    \includegraphics[width=0.95\linewidth]{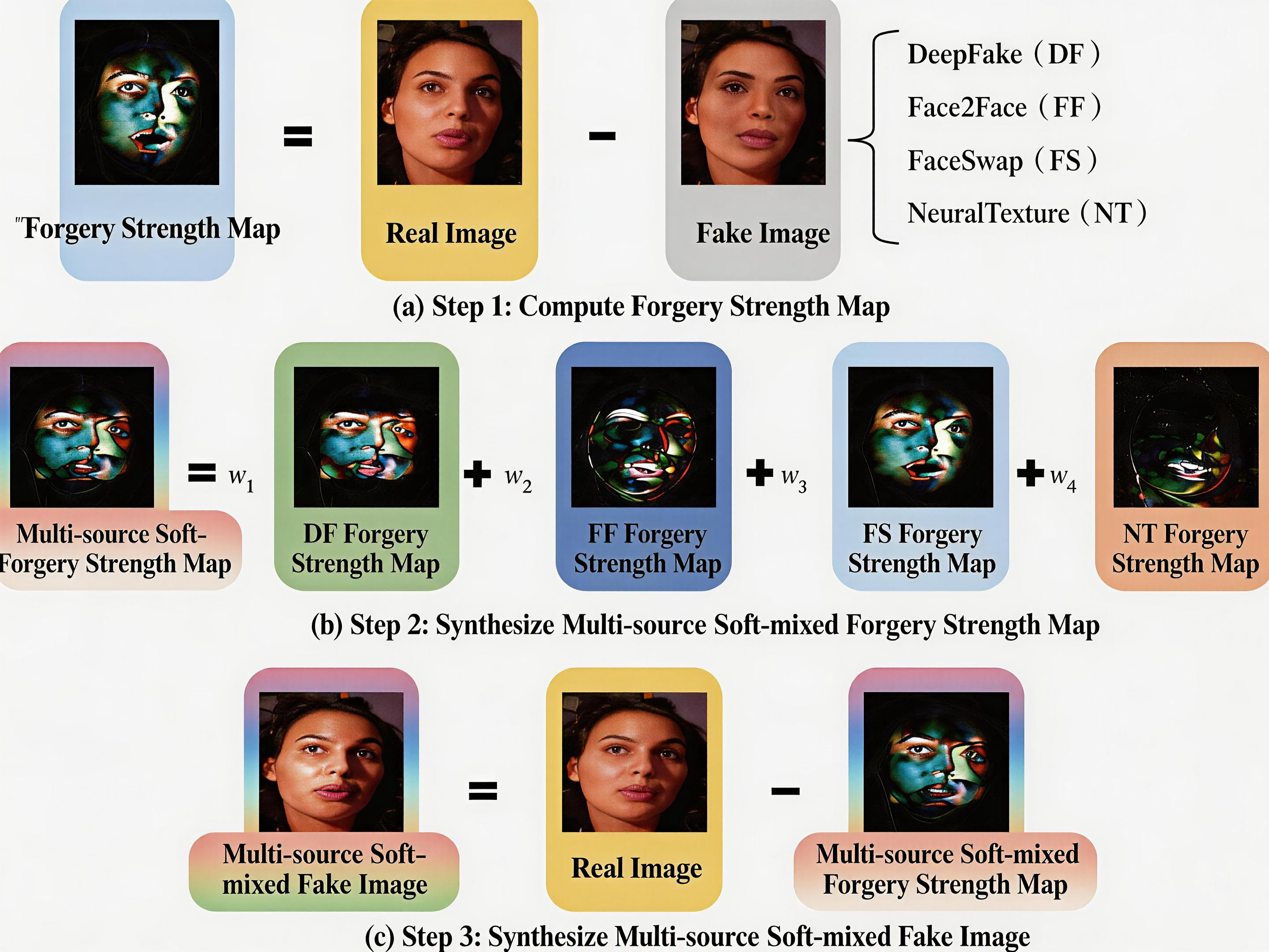} % Replace with actual figure path
    \caption{Illustration of the Multivariate and Soft Blending Augmentation (MSBA) process. (a) Step 1: Calculate per-method forgery intensity maps ($M_i$) between the real image and each forged version. (b) Step 2: Synthesize a blended intensity map $\tilde{M}$ by linearly combining individual maps with random weights $\alpha_i$. (c) Step 3: Generate the final blended forgery image $\tilde{I}$ by subtracting the scaled blended map from the real image.}
    \label{fig:msba}
\end{figure}

Let $I_{\text{real}}$ denote a pristine face image. From the training set, we have corresponding forged versions using $M$ different methods: $\{I_{\text{forge}}^{(1)}, I_{\text{forge}}^{(2)}, ..., I_{\text{forge}}^{(M)}\}$ (e.g., $M=4$ for FF++). For each forged image, we first compute a pixel-wise \textit{forgery intensity map} $M_i$ by taking the absolute difference between the forged image and the real image in a normalized color space:
\begin{equation}
    M_i = | I_{\text{real}} - I_{\text{forge}}^{(i)} |, \quad i \in \{1,...,M\}
\end{equation}
These maps highlight regions altered by each specific forgery technique, as visualized in Figure \ref{fig:msba}(a).

The MSBA process then synthesizes a new image $\tilde{I}$ and its corresponding blended intensity map $\tilde{M}$ by convexly combining the individual forgery maps with random weights $\alpha_i$:
\begin{equation}
    \tilde{M} = \sum_{i=1}^{M} \alpha_i \cdot M_i, \quad \text{where } \alpha_i \sim \text{Dirichlet}(\boldsymbol{\beta}), \sum_{i=1}^{M} \alpha_i = 1
\end{equation}
\begin{equation}
    \tilde{I} = I_{\text{real}} - \lambda \cdot \tilde{M}
\end{equation}
Here, $\lambda$ is a scaling factor controlling the overall intensity of the forgery blend, and $\text{Dirichlet}(\boldsymbol{\beta})$ is the Dirichlet distribution which ensures the blending weights $\{\alpha_i\}$ sum to 1, promoting a balanced mixture of forgery traces. The synthesized image $\tilde{I}$ (Figure \ref{fig:msba}(c)) and its corresponding \textit{soft label} $\boldsymbol{\alpha} = [\alpha_1, ..., \alpha_M]$ are used for training. This process compels the network to learn to recognize and disentangle multiple, overlapping forgery patterns simultaneously, significantly enhancing its robustness against unseen or hybrid forgery techniques.

\subsection{Multivariate Forgery Intensity Estimation Module}
\label{subsec:mfie}

\begin{figure}[t]
    \centering
    \includegraphics[width=0.8\linewidth]{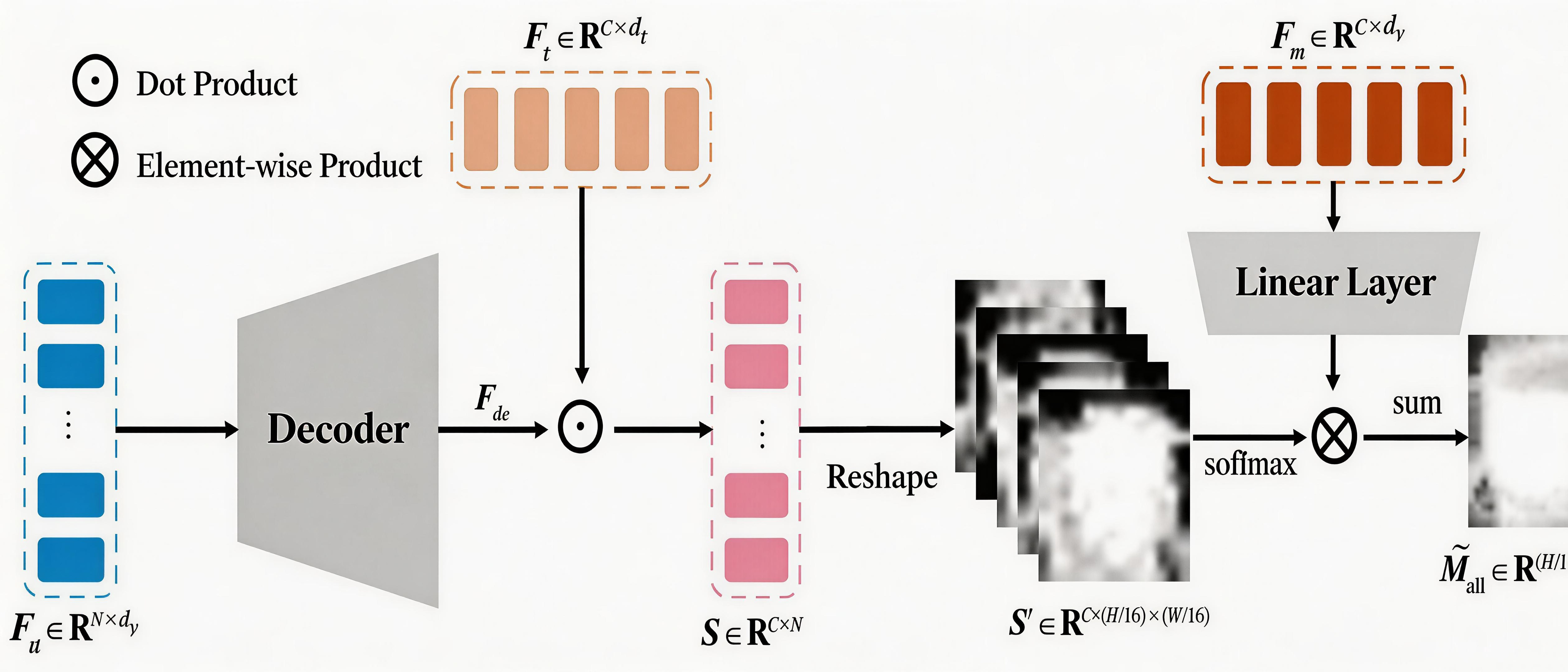} % Replace with actual figure path
    \caption{Detailed architecture of the Multivariate Forgery Intensity Estimation (MFIE) module. It takes patch-level features from the visual encoder, upsamples them, and estimates both a per-pixel forgery intensity map and the method-wise blending weights.}
    \label{fig:mfi_module}
\end{figure}

Recognizing that different forgery methods and intensities produce artifacts of varying strengths and characteristics, we design a \textbf{Multivariate Forgery Intensity Estimation (MFIE)} module. This module, illustrated in Figure \ref{fig:mfi_module}, operates in parallel with the main classification pathway and has two objectives: 1) to estimate the overall forgery intensity at each spatial location, and 2) to predict the blending weights $\boldsymbol{\alpha}$ used in the MSBA process, providing an auxiliary self-supervision signal.

Let $F_p \in \mathbb{R}^{N \times d_v}$ denote the patch-level features extracted from the penultimate layer of the visual encoder (excluding the [CLS] token). These features are processed by a lightweight decoder consisting of two transposed convolution layers to upsample the spatial resolution, producing decoded features $F_{de} \in \mathbb{R}^{H' \times W' \times d_v}$.

\textbf{Forgery Intensity Map Prediction:} The decoder features $F_{de}$ are projected to a channel dimension $C$, and a per-pixel softmax is applied to obtain a channel-wise probability distribution $\tilde{M}_i$ over $C$ potential forgery patterns. The final predicted intensity map $\tilde{M}_{\text{all}}$ is a weighted sum of these channel maps, where the weights $\tilde{w}_i$ are predicted from the global [CLS] token via a linear layer, as shown in Figure \ref{fig:mfi_module}.

\textbf{Blending Weight Prediction:} The global [CLS] token representation $F_{\text{cls}}$ is also fed into a separate linear layer to predict the estimated blending weights $\hat{\boldsymbol{\alpha}} = [\hat{\alpha}_1, ..., \hat{\alpha}_M]$. This prediction is supervised against the ground-truth soft label $\boldsymbol{\alpha}$ used during MSBA.

The MFIE module serves as an explicit regularizer, guiding the visual encoder to develop a fine-grained understanding of forgery intensity and composition, which is complementary to the primary binary classification objective.

\subsection{Multi-Task Learning Objectives}
\label{subsec:loss}

Our model is trained with a composite loss function $\mathcal{L}_{\text{total}}$ that combines the main classification loss with auxiliary losses from the MFIE module and the semantic similarity objective:
\begin{equation}
    \mathcal{L}_{\text{total}} = \lambda_{\text{cls}} \mathcal{L}_{\text{cls}} + \lambda_{\text{sim}} \mathcal{L}_{\text{sim}} + \lambda_{\text{int}} \mathcal{L}_{\text{int}} + \lambda_{\text{wgt}} \mathcal{L}_{\text{wgt}}
\end{equation}

\noindent where:
\begin{itemize}
    \item \textbf{Binary Classification Loss ($\mathcal{L}_{\text{cls}}$)}: Standard binary cross-entropy loss applied to the final fused prediction $\hat{y}$ against the ground-truth label $y \in \{0,1\}$.
    \item \textbf{Semantic Similarity Loss ($\mathcal{L}_{\text{sim}}$)}: Binary cross-entropy loss applied to the similarity score $\tilde{y}_{\text{sim}}$ against the ground-truth label $y$. This encourages the image features to be semantically aligned with high-level text descriptions of forgeries.
    \item \textbf{Forgery Intensity Estimation Loss ($\mathcal{L}_{\text{int}}$)}: Smooth L1 loss between the predicted intensity map $\tilde{M}_{\text{all}}$ and the ground-truth blended intensity map $\tilde{M}$ (resized and quantized to patch-level):
    \begin{equation}
        \mathcal{L}_{\text{int}} = \text{SmoothL1}(\tilde{M}_{\text{all}}, \text{Resize}(\tilde{M}))
    \end{equation}
    \item \textbf{Blending Weight Prediction Loss ($\mathcal{L}_{\text{wgt}}$)}: Kullback–Leibler (KL) divergence loss between the predicted blending weight distribution $\hat{\boldsymbol{\alpha}}$ and the true Dirichlet distribution parameters $\boldsymbol{\alpha}$:
    \begin{equation}
        \mathcal{L}_{\text{wgt}} = D_{\text{KL}}(\boldsymbol{\alpha} \, \|\, \hat{\boldsymbol{\alpha}})
    \end{equation}
\end{itemize}

The hyperparameters $\lambda_{\text{cls}}, \lambda_{\text{sim}}, \lambda_{\text{int}}, \lambda_{\text{wgt}}$ control the balance between these objectives. Through this multi-task learning paradigm, the model jointly optimizes for accurate forgery detection, semantic alignment, and detailed forgery pattern understanding, leading to a more robust and generalizable representation.


\section{Experiments}
\label{sec:experiments}

In this section, we present a comprehensive evaluation of the proposed MSBA-CLIP framework. We begin by detailing the experimental setup, including datasets, evaluation metrics, and implementation specifics. Subsequently, we report both in-domain and cross-domain performance, comparing our method against several state-of-the-art baselines. Robustness tests under various perturbations and extensive ablation studies are conducted to validate the contribution of each component. Finally, we provide visualizations and a performance analysis to offer deeper insights.

\subsection{Datasets and Evaluation Metrics}
\label{subsec:datasets_metrics}

\subsubsection{Datasets}
Our experiments utilize one primary training dataset and five independent testing datasets to thoroughly assess generalization capability.

\textbf{Training Dataset:} We employ the FaceForensics++ (FF++) dataset for training. FF++ consists of 1,000 original real videos collected from YouTube and 4,000 corresponding fake videos generated using four distinct face-swapping techniques: DeepFakes (DF), FaceSwap (FS), Face2Face (F2F), and NeuralTextures (NT). The dataset provides videos in three compression qualities: raw (lossless), High Quality (C23), and Low Quality (C40). In line with standard practice, we adopt the C23 version for our main experiments. The dataset is partitioned into training, validation, and test sets with an 8:1:1 ratio.

\textbf{Testing Datasets:} To evaluate generalization, we test on five widely-used, independent deepfake detection benchmarks:
\begin{itemize}
    \item \textbf{Celeb-DF (CDF) v2} \cite{li2024longcontextllmsstrugglelong}: A high-quality dataset featuring 590 real and 5,639 fake videos of celebrities, known for its visual realism.
    \item \textbf{DeepFake Detection Challenge (DFDC) Preview} \cite{wang2025fine}: A diverse preview set from the DFDC competition.
    \item \textbf{DeepFake Detection Challenge (DFDC)} \cite{es2023ragasautomatedevaluationretrieval}: The full, large-scale dataset from the DFDC, containing over 100,000 clips.
    \item \textbf{DeepFake Detection (DFD)} \cite{shan2024mctbench}: Google's DeepFakeDetection dataset.
    \item \textbf{DeeperForensics-1.0 (DFo)} \cite{fu2024ocrbench}: A challenging dataset designed with improved anti-forensic perturbations.
\end{itemize}
This suite of datasets covers a broad spectrum of generation methods, source identities, and video qualities, providing a rigorous test for model robustness.

\subsubsection{Evaluation Metrics}
Following established protocols in the field, we report frame-level performance using two key metrics:
\begin{itemize}
    \item \textbf{Accuracy (ACC)}: The proportion of correctly classified frames (real or fake) among all frames.
    \item \textbf{Area Under the Receiver Operating Characteristic Curve (AUC)}: A threshold-independent metric that measures the model's ability to distinguish between real and fake samples across all possible decision thresholds. A higher AUC indicates better overall discriminative power.
\end{itemize}
For a more holistic evaluation on cross-dataset benchmarks, we also compute video-level AUC by averaging the prediction scores of all frames within a video.

\subsection{Implementation Details}
\label{subsec:implementation}

\textbf{Data Preprocessing:} For training on FF++, we uniformly sample 16 frames from each video. For testing across all datasets, we sample 32 frames per video to ensure a stable video-level prediction. All frames are resized to $224 \times 224$ pixels and normalized using ImageNet statistics.

\textbf{Model Configuration:} Our framework is built upon the CLIP-ViT/B-16 model \cite{zhang2023blind}, which is pre-trained on a large-scale web-collected image-text dataset. The text prompts for the four forgery types are constructed as described in Section \ref{subsec:overall_arch}. For the $L$ generic fake descriptions used in the semantic similarity loss, we use $L=16$ phrases such as "A manipulated face", "This is a synthetic portrait", etc. The Multimodal Interaction Projection (MIP) layer is a two-layer MLP with a hidden dimension of $512$ and GELU activation.

\textbf{Training Protocol:} We use the Adan optimizer \cite{Xu2024} with an initial learning rate of $2 \times 10^{-5}$ and a batch size of 64. The model is trained for 75 epochs, followed by 25 epochs of fine-tuning with a cosine annealing scheduler that decays the learning rate to $2 \times 10^{-7}$. The loss weights are empirically set to $\lambda_{\text{cls}}=1.0$, $\lambda_{\text{sim}}=0.5$, $\lambda_{\text{int}}=1.0$, and $\lambda_{\text{wgt}}=0.1$. All experiments are conducted on a single NVIDIA GeForce RTX 3090 GPU using PyTorch.

\textbf{MSBA Parameters:} During training, the blending weights $\alpha_i$ for the four forgery methods are sampled from a symmetric Dirichlet distribution with concentration parameter $\beta = 1.0$. The intensity scaling factor $\lambda$ is randomly chosen from $[0.8, 1.2]$ for each sample to increase diversity.

\subsection{Comparison with State-of-the-Art Methods}
\label{subsec:sota_comparison}

We compare our MSBA-CLIP method against a wide range of recent and competitive deepfake detection approaches. The baselines include: \textbf{Xception}(a CNN-based baseline), \textbf{Face X-ray}  (blending artifact detection), \textbf{F$^3$Net} (frequency-aware forgery detection), \textbf{SPSL}  (spatial-phase shallow learning), \textbf{SRM}  (noise pattern analysis with SRM filters), \textbf{UCF}  (multi-task common feature learning), and \textbf{CORE} .

\subsubsection{In-Domain Evaluation}
\label{subsubsec:in_domain}
We first evaluate the model's performance on the FF++ dataset under both high-quality (C23) and low-quality (C40) settings, where training and testing data share the same source distribution. The results are presented in Table \ref{tab:in_domain}.

\begin{table}[htbp]
\centering
\caption{In-domain evaluation results on the FF++ dataset. Bold indicates the best performance.}
\label{tab:in_domain}
\begin{tabular}{lcccc}
\toprule
\multirow{2}{*}{Method} & \multicolumn{2}{c}{FF++ (C40)} & \multicolumn{2}{c}{FF++ (C23)} \\
\cmidrule(lr){2-3} \cmidrule(lr){4-5}
 & ACC (\%) & AUC (\%) & ACC (\%) & AUC (\%) \\
\midrule
Xception  & 61.60 & 85.59 & 87.35 & 98.70 \\
Face X-ray & 81.75 & 95.73 & -- & -- \\
F$^3$Net  & 88.69 & 90.40 & 97.60 & 99.29 \\
SPSL  & 92.89 & 95.31 & 97.93 & 99.51 \\
SRM  & -- & -- & 94.48 & 98.91 \\
UCF  & 96.68 & 97.47 & 99.68 & 99.99 \\
\textbf{MSBA-CLIP (Ours)} & \textbf{100.00} & \textbf{100.00} & \textbf{100.00} & \textbf{100.00} \\
\bottomrule
\end{tabular}
\end{table}

As shown, our method achieves perfect scores (100\% ACC and AUC) on both C23 and C40 test splits, significantly outperforming all baselines. This demonstrates that the proposed framework, through its synergistic use of multimodal alignment (CLIP) and sophisticated augmentation (MSBA), learns highly discriminative and robust features that generalize perfectly within the FF++ domain, even under the challenging heavy compression of C40.

\subsubsection{Cross-Domain Evaluation}
\label{subsubsec:cross_domain}
The core challenge in deepfake detection is generalization to unseen manipulation techniques. We train our model exclusively on FF++ (C23) and evaluate its performance on five independent datasets. Table \ref{tab:cross_domain_frame} presents the frame-level AUC results.

\begin{table}[htbp]
\centering
\caption{Cross-domain evaluation (frame-level AUC \%) on five independent datasets. Models are trained on FF++ (C23). Bold indicates the best performance.}
\label{tab:cross_domain_frame}
\begin{tabular}{lccccc|c}
\toprule
Method & CDF & DFDC & DFDCP & DFo & DFD & \textbf{Avg.} \\
\midrule
Xception & 73.65 & 70.77 & 73.74 & 83.30 & 81.63 & 76.62 \\
Face X-ray& 79.70 & 65.50 & -- & 86.80 & -- & -- \\
F$^3$Net & 73.52 & 70.21 & 73.54 & 84.31 & 79.75 & 76.27 \\
SPSL & 76.50 & 70.40 & 74.08 & 87.67 & 81.22 & 77.97 \\
SRM & 75.52 & 69.95 & 74.08 & 86.38 & 81.20 & 77.43 \\
UCF & 75.27 & 71.91 & 75.94 & 82.41 & 80.74 & 77.25 \\
CORE & 74.28 & 70.49 & 73.41 & 84.75 & 80.18 & 76.62 \\
 \textbf{MSBA-CLIP (Ours)} & \textbf{78.18} & \textbf{72.98} & \textbf{75.56} & \textbf{88.55} & \textbf{90.93} & \textbf{81.24} \\
\bottomrule
\end{tabular}
\end{table}

Our method consistently achieves the highest AUC on all five cross-domain datasets, with an average improvement of \textbf{3.27\%} over the strongest baseline (SPSL). The most notable gain is observed on the DFD dataset (+9.73\% over UCF), which contains different source identities and generation pipelines not seen during training. This underscores the exceptional generalization capability endowed by our multimodal and augmentation strategies.

Video-level results, obtained by averaging frame scores per video, are reported in Table \ref{tab:cross_domain_video}. Our method maintains superior performance, particularly on CDF and DFD, further confirming its robustness for practical video-based detection scenarios.

\begin{table}[htbp]
\centering
\caption{Cross-domain evaluation (video-level AUC \%) on selected datasets. Bold indicates the best performance.}
\label{tab:cross_domain_video}
\begin{tabular}{lcccc}
\toprule
Method & CDF & DFDC & DFDCP & DFD \\
\midrule
Face X-ray & 80.58 & -- & 80.92 & 95.40 \\
CORE & 75.71 & -- & 71.41 & 94.09 \\
UCF & 83.70 & 71.40 & 80.51 & 93.10 \\
IID & 83.80 & -- & 81.23 & 93.92 \\
 \textbf{MSBA-CLIP (Ours)} & \textbf{86.10} & \textbf{76.37} & \textbf{80.38} & \textbf{97.19} \\
\bottomrule
\end{tabular}
\end{table}

\subsection{Robustness Analysis}
\label{subsec:robustness}
\begin{figure}[htbp]
\centering
\includegraphics[width=0.85\linewidth]{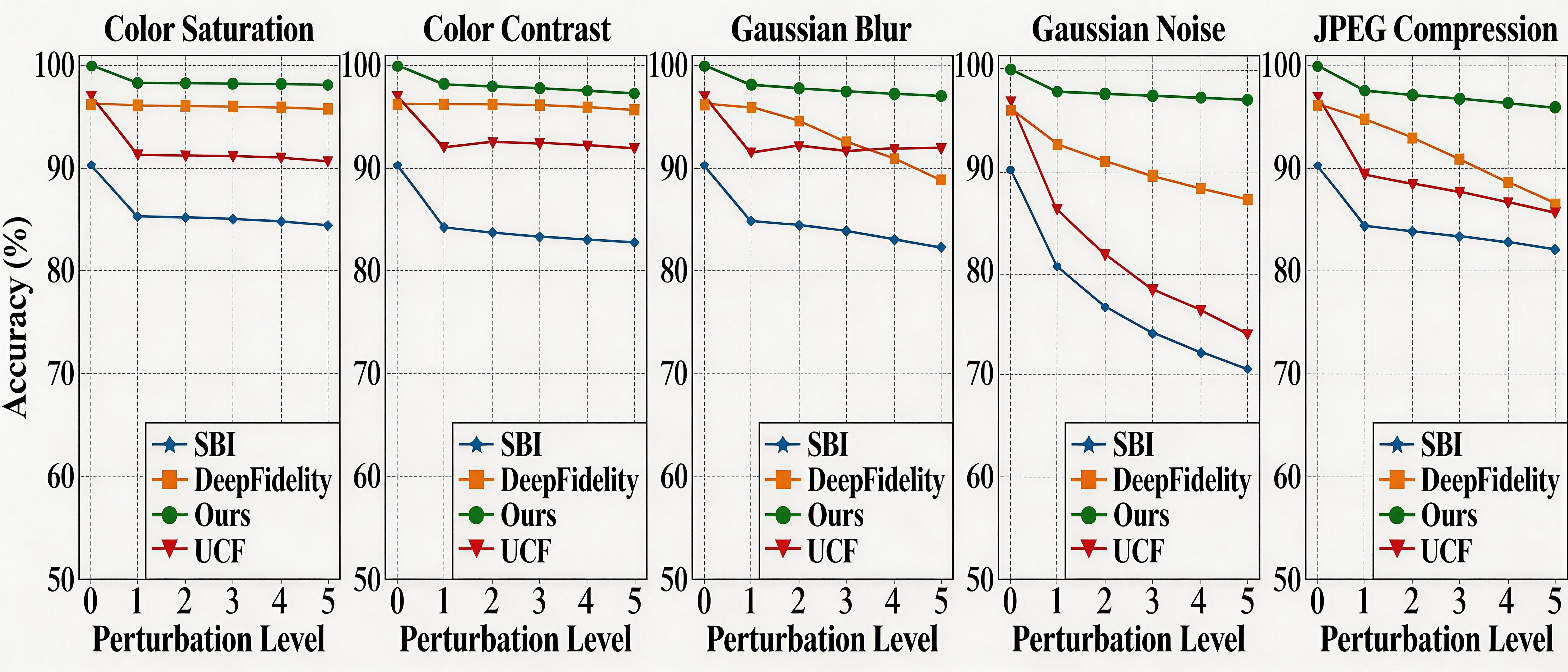}
\caption{Robustness evaluation under five types of image perturbations at varying intensity levels. Our method (MSBA-CLIP) demonstrates superior stability compared to state-of-the-art baselines.}
\label{fig:robustness}
\end{figure}

Real-world applications often involve videos subjected to various post-processing or compression. To evaluate robustness, we apply five common perturbations—Gaussian Blur, Gaussian Noise, JPEG Compression, Color Saturation Change, and Color Contrast Change—to the FF++ (C23) test set, each at five intensity levels. We compare our method against three recent state-of-the-art methods: DeepFidelity \cite{paddleocr2023}, SBI \cite{shan2024mctbench}, and UCF \cite{yu2025benchmarking}. The results, depicted in Figure \ref{fig:robustness}, show that our method exhibits the smallest performance degradation across all perturbation types, especially under JPEG compression and noise addition. This robustness is attributed to the CLIP backbone's pre-training on diverse, noisy internet data and the MSBA strategy's encouragement of learning intrinsic forgery semantics rather than surface-level artifacts.

\subsection{Ablation Studies}
\label{subsec:ablation}
We conduct extensive ablation studies to dissect the contribution of each proposed component. All models are trained on FF++ (C23) and evaluated on the five cross-domain datasets.

\subsubsection{Effectiveness of Individual Modules}
Table \ref{tab:ablation_modules} presents the results of progressively adding our proposed modules to a CLIP baseline. The baseline uses only the CLIP image encoder with a standard classification head.

\begin{table}[htbp]
\centering
\caption{Ablation study on the contribution of each proposed module. Average frame-level AUC (\%) across five cross-domain datasets is reported.}
\label{tab:ablation_modules}
\begin{tabular}{lccccc|c}
\toprule
MSBA & MFIE & CDF & DFDC & DFo & DFD & \textbf{Avg.} \\
\midrule
-- & -- & 73.85 & 71.89 & 72.11 & 87.49 & 81.42 \\
 $\checkmark$ & -- & 75.12 & 73.05 & 74.50 & 89.25 & 83.71 \\
 $\checkmark$ & $\checkmark$ & \textbf{78.18} & \textbf{72.98} & \textbf{75.56} & \textbf{90.93} & \textbf{86.84} \\
\bottomrule
\end{tabular}
\end{table}

\textbf{Adding MSBA}: Incorporating the Multivariate and Soft Blending Augmentation strategy alone boosts the average AUC by +2.29\%. This confirms that training on blended forgeries forces the model to learn more generalizable features, reducing overfitting to method-specific artifacts.

\textbf{Adding MFIE}: Further integrating the Multivariate Forgery Intensity Estimation module leads to an additional +3.13\% average gain. The MFIE module provides fine-grained supervision, guiding the model to attend to and quantify subtle forgery traces, which is particularly beneficial for detecting low-intensity or complex blended manipulations.

\subsubsection{Impact of the Number of Text Prompts ($N$)}

Our framework uses $N$ generic negative text descriptions for the semantic similarity loss. Table \ref{tab:ablation_n} shows that performance plateaus and is optimal when $N=16$. Using too few prompts ($N=1$ or $4$) limits textual diversity, while an excessive number ($N>16$) may introduce noise or over-regularization.

\begin{table}[htbp]
\centering
\caption{Ablation on the number of generic negative text prompts ($N$) used in the semantic similarity loss.}
\label{tab:ablation_n}
\begin{tabular}{cccccc|c}
\toprule
$N$ & CDF & DFDC & DFo & DFD & \textbf{Avg.} \\
\midrule
1 & 73.07 & 75.03 & 78.16 & 84.67 & 84.58 \\
4 & 72.70 & 73.04 & 74.40 & 83.87 & 82.31 \\
8 & 74.19 & 76.19 & 79.19 & 86.63 & 84.04 \\
 16 & \textbf{78.18} & \textbf{76.37} & \textbf{80.38} & \textbf{90.93} & \textbf{86.84} \\
\bottomrule
\end{tabular}
\end{table}

\subsubsection{Analysis of Loss Weight Coefficients}
The balance between the four loss components is crucial. We experiment with different weight combinations for $\lambda_{\text{sim}}$ (similarity loss) and $\lambda_{\text{int}}$ (intensity loss), keeping $\lambda_{\text{cls}}=1$ and $\lambda_{\text{wgt}}=0.1$. Results in Table \ref{tab:ablation_loss} indicate that equal weighting ($\lambda_{\text{sim}}=1, \lambda_{\text{int}}=1$) yields the best overall performance, suggesting that the classification, semantic alignment, and intensity estimation objectives are complementary and of comparable importance.

\begin{table}[htbp]
\centering
\caption{Ablation on the weights of the auxiliary loss terms. $\lambda_{\text{cls}}$ is fixed to 1.0.}
\label{tab:ablation_loss}
\begin{tabular}{ccc|cccc|c}
\toprule
$\lambda_{\text{sim}}$ & $\lambda_{\text{int}}$ & $\lambda_{\text{wgt}}$ & CDF & DFDC & DFo & DFD & \textbf{Avg.} \\
\midrule
 1 & 1 & 0.1 & \textbf{78.18} & \textbf{76.37} & \textbf{80.38} & \textbf{90.93} & \textbf{86.84} \\
1 & 10 & 0.1 & 75.02 & 71.09 & 73.96 & 81.43 & 80.97 \\
10 & 1 & 0.1 & 76.76 & 77.35 & 78.93 & 85.87 & 82.05 \\
10 & 10 & 0.1 & 77.51 & 75.72 & 79.20 & 82.22 & 83.69 \\
\bottomrule
\end{tabular}
\end{table}

\subsubsection{Computational Cost and Performance Trade-off}
We compare the efficiency of our model against three recent efficient models in Table \ref{tab:performance}. While our model has more parameters due to the CLIP backbone, it achieves a superior trade-off, offering the highest AUC with moderate FLOPs, demonstrating its practical viability.

\begin{table}[htbp]
\centering
\caption{Performance and efficiency comparison. $\uparrow$ indicates higher is better, $\downarrow$ indicates lower is better.}
\label{tab:performance}
\begin{tabular}{lccc}
\toprule
Method & FLOPs (G) $\downarrow$ & Params (M) $\downarrow$ & AUC (\%) $\uparrow$ \\
\midrule
RECCE  & 8.09 & 23.78 & 73.19 \\
SRM  & 13.81 & 53.24 & 75.52 \\
UCF  & 12.19 & 44.51 & 75.27 \\
 \textbf{MSBA-CLIP (Ours)} & 11.55 & 57.65 & \textbf{78.18} \\
\bottomrule
\end{tabular}
\end{table}

\subsection{Qualitative Analysis and Visualization}
\label{subsec:visualization}
\begin{figure}[htbp]
    \centering
    \includegraphics[width=0.95\linewidth]{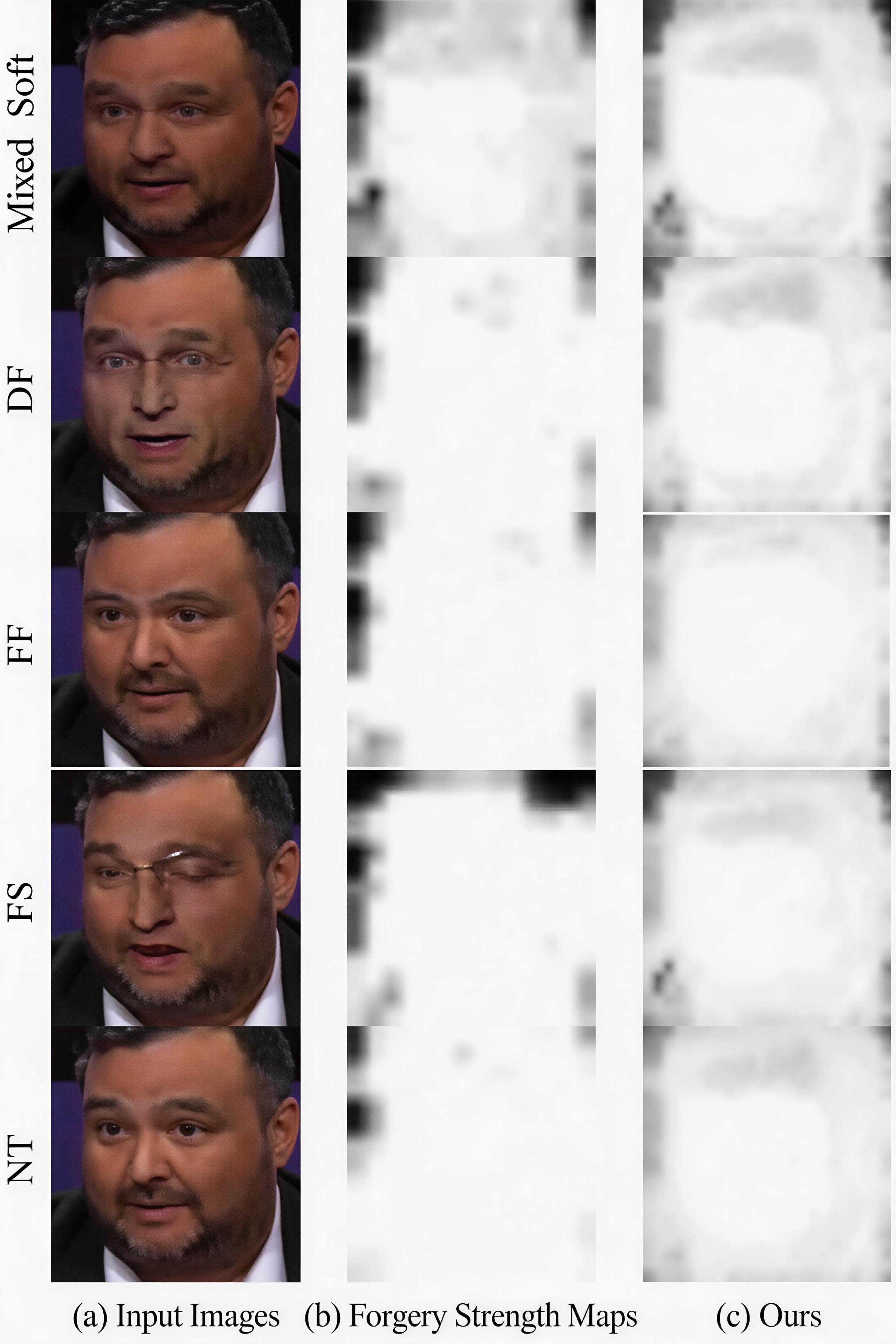}
    \caption{Visualization of forgery intensity maps. (a) Input images (Real, Blended, DF, FF, FS, NT). (b) Ground-truth intensity maps (patch-level, multiplied by 5 for visibility). (c) Predicted intensity maps from our MFIE module. The model accurately localizes manipulated regions across various forgery types and complex blends.}
    \label{fig:visualization}
\end{figure}

To gain deeper insight into the model's decision-making process, we visualize the predicted forgery intensity maps from the MFIE module. Figure \ref{fig:visualization} compares the ground-truth patch-level intensity maps with the predictions for images from different forgery methods (DF, FF, FS, NT) and an MSBA-synthesized blended image.

The model successfully localizes manipulated facial regions (e.g., mouth, eyes, cheek contours) with high precision. Notably, for the MSBA-blended image, the predicted map reflects a composite of intensities corresponding to the contributing forgery methods, demonstrating the module's ability to disentangle and estimate multivariate forgery signals. These visualizations confirm that our model learns to focus on semantically meaningful forgery traces rather than dataset-specific biases.

\section{Conclusion}
This paper presents a novel wavelet-suppressed diffusion model for dual-channel blind image separation. The proposed DCDSM framework effectively addresses complex mixture separation challenges through interactive dual-branch processing and wavelet-frequency domain feature extraction. Extensive experiments demonstrate state-of-the-art performance in rain/snow removal and complex mixture separation tasks. Future work will focus on computational efficiency improvement through latent diffusion models and extension to emerging applications like document image restoration.

\clearpage

\nocite{*}
\bibliographystyle{IEEEtran}
\bibliography{custom}

\end{document}